%% file: main.tex
\documentclass[conference]{IEEEtran}

\makeatletter

\def\ps@IEEEtitlepagestyle{%
  \def\@oddfoot{}%
  \def\@evenfoot{}%
}
\def\mycopyrightnotice{%
  {\footnotesize XXX-X-XXXX-XXXX-X/XX/\$XX.00~\copyright~20XX IEEE\hfill}
  \gdef\mycopyrightnotice{}
}

\usepackage{blindtext}
\usepackage{eso-pic}
\IEEEoverridecommandlockouts
\usepackage{cite}
\usepackage{amsmath,amssymb,amsfonts}
\usepackage{algorithmic}
\usepackage{graphicx}
\usepackage{textcomp}
\usepackage{xcolor}
\usepackage{graphicx}
\usepackage{xurl}
\PassOptionsToPackage{hyphens}{url}\usepackage{hyperref}
\usepackage{float}
\usepackage{subfig}
\usepackage{caption}
\usepackage{amsthm}
\usepackage{algorithm}
\usepackage{booktabs}

\def\BibTeX{{\rm B\kern-.05em{\sc i\kern-.025em b}\kern-.08em
    T\kern-.1667em\lower.7ex\hbox{E}\kern-.125emX}}
    
\usepackage{eso-pic}
\newcommand\AtPageUpperMyright[1]{\AtPageUpperLeft{%
 \put(\LenToUnit{0.17\paperwidth},\LenToUnit{-2cm}){%
     \parbox{0.9\textwidth}{\raggedleft\fontsize{8}{11}\selectfont #1}}%
 }}%
\newcommand{\conf}[1]{%
\AddToShipoutPictureBG*{%
\AtPageUpperMyright{#1}
}
}

\newcommand{\orcid}[1]{\href{https://orcid.org/#1}{#1}}
    
\begin{document}
\title{\vspace*{1cm} Improving Line Search Methods for Large Scale Neural Network Training
\thanks{The authors were supported by SAIL. 
SAIL is funded by the Ministry of Culture and Science of the State of North Rhine-Westphalia under the grant no NW21-059A.}
}

\author{\IEEEauthorblockN{ Philip Kenneweg}
\IEEEauthorblockA{\textit{Technical Faculty} \\
\textit{University Bielefeld}\\
Bielefeld, Germany \\
\orcid{0000-0002-7097-173X}}
\and
\IEEEauthorblockN{ Tristan Kenneweg}
\IEEEauthorblockA{\textit{Technical Faculty} \\
\textit{University Bielefeld}\\
Bielefeld, Germany \\
\orcid{0000-0001-8213-9396}}
\and
\IEEEauthorblockN{  Barbara Hammer}
\IEEEauthorblockA{\textit{Technical Faculty} \\
\textit{University Bielefeld}\\
Bielefeld, Germany \\
 \orcid{0000-0002-0935-5591}}
}

\maketitle
\conf{\textit{  Proc. of International Conference on Artificial Intelligence, Computer, Data Sciences and Applications (ACDSA 2024) \\ 
1-2 February 2024, Victoria-Seychelles}}
\begin{abstract}
In recent studies, line search methods have shown significant improvements in the performance of traditional stochastic gradient descent techniques, eliminating the need for a specific learning rate schedule. 
In this paper, we identify existing issues in state-of-the-art line search methods, 
propose enhancements, and rigorously evaluate their effectiveness. We test these methods on larger datasets and more complex data domains than before.

Specifically, we improve the Armijo line search by integrating the momentum term from ADAM in its search direction, enabling efficient large-scale training, a task that was previously prone to failure using Armijo line search methods. Our optimization approach outperforms both the previous Armijo implementation and tuned learning rate schedules for Adam.

Our evaluation focuses on Transformers and CNNs in the domains of NLP and image data.

Our work is publicly available as a Python package, which provides a hyperparameter free Pytorch optimizer.
\end{abstract}



\section{Introduction}
\label{sec:intro}

\input{intro}

\section{Background}
\label{sec:Background}
\input{background}

\section{Methods}
\label{sec:methods}
\input{methods}
\section{Experimental Design}
\label{sec:experiments}
\input{experiments}

\section{Results}
\label{sec:results}

\input{results}



\section{Conclusion}
\label{sec:conclusion}
\input{conclusion}

\bibliography{references}
\bibliographystyle{ieeetr}

\end{document}

%% file: intro.tex
In the field of modern machine learning, there is a wide array of optimization algorithms available, some examples are SGD \cite{robbins51a}, RADAM \cite{DBLP:journals/corr/abs-1908-03265}, AdamW \cite{adamw}, RMSprop \cite{RMSprop} and Adam \cite{adam}. 
Nevertheless, selecting the most appropriate algorithm for a specific problem and determining the right learning rate or learning rate schedule often demands considerable expertise and computational resources \cite{schmidt2021descending}. Typically, this involves treating the learning rate as a hyperparameter and repeatedly training the network until the optimal value that maximizes performance is found.

Recent research in deep learning \cite{vaswani20a, mahsereci15a, bollapragada18a, paquette20a} has suggested the resurgence of line search methods as a prominent optimization technique. These methods efficiently determine an adaptive learning rate by evaluating the loss function at various points along the gradient direction. This approach eliminates the need for laborious hyperparameter tuning and can yield superior performance compared to manually set learning rates.

Classical line search methods necessitate multiple forward passes for each update step, which can be computationally expensive. To address this, \cite{vaswani20a} introduced a more efficient approach called Stochastic Line Search (SLS) coupled with an intelligent step size re-initialization. This method, as demonstrated in their study, enhances the performance of various optimization techniques, including Stochastic Gradient Descent (SGD), for tasks like matrix factorization and image classification with small networks and datasets. Moreover, in a subsequent work \cite{vaswani2021adaptive}, this line search technique was adapted for use with preconditioned optimizers such as Adam, further extending its applicability.

In this paper, we extend upon this work and previous research by us \cite{ijcnn2023}, by integrating the momentum term from ADAM, resulting in performance and stability improvements. Furthermore, we perform comprehensive experiments to assess various optimization techniques across diverse datasets and architectural choices. Our results consistently show that our enhanced Automated Large Scale ADAM Line Search (ALSALS) algorithm outperforms both the previously introduced SLS and fine-tuned optimizers.

To enhance the replicability and accessibility of our work, we have implemented all methods as PyTorch optimizers. The source code is openly available as free software under the MIT license and can be accessed at \url{https://github.com/TheMody/Improving-Line-Search-Methods-for-Large-Scale-Neural-Network-Training}

%% file: background.tex

The stochastic Armijo line search, as detailed in \cite{vaswani20a} and further elaborated upon in \cite{ijcnn2023}, aims to establish an appropriate step size for all network parameters $w_k$ at iteration $k$. in this section, we formulate a modification of the Armijo criterion to handle the ADAM \cite{adam} direction instead of the classical SGD direction. This is based upon \cite{vaswani20a, vaswani2021adaptive}
Moreover, we introduce an improved Armijo criterion, which mitigates the effect of noise in the mini-batch setting by calculating an exponential moving average smoothing on both sides of the Armijo equation.

We use common notations from previous papers, see \cite{ijcnn2023}.
\subsection{Armijo Line Search} 
\label{sec:sgdarmijo}
The Armijo line search criterion is defined in \cite{vaswani20a} as:
\begin{equation}
    f_{k}(w_k + \eta_k d_k) \leq f_{k}(w_k) - c \cdot \eta_k ||\nabla f_{k}(w_k)||^2,
    \label{eq:armijo}
\end{equation}
where $d_k$ is the direction (e.g., $d_k=-\nabla f_{k}(w_k)$ in case of SGD), 
$c \in (0,1)$ is a hyper-parameter which regulates the step size (in other work set to $0.1$ \cite{vaswani20a}). The step size $\eta_k$ which satisfies Condition \ref{eq:armijo} is obtained by performing a backtracking procedure, see \cite{armijo66a}. 

To enable step size growth, $\eta_k$ is increased each step by the following formula:
\begin{equation}\label{eq:tweak}
	\eta^0_k = \eta_{k-1} \cdot 2^{1/b}
\end{equation}
as described in \cite{vaswani20a}. In practice for $b=500$, this will usually avoid backtracking multiple times per step, since the increase in step size is small. Henceforth, we will refer to this algorithm as SLS.

\subsection{Including Adam's Update Step in SLS}


\label{sec:adamarmijo}
In the case of Stochastic Gradient Descent, the direction of descent $d_k$ is the negative mini-batch gradient e.g.
\begin{equation*}
     d_k = -\nabla f_{k}(w_k)
\end{equation*}
Adam's descent direction and magnitude $d_k$  defined in \cite{adam} can be written as:
\begin{equation}
\begin{split}
   g_{k} &= \nabla f_{ik}(w_k) \\
   m_{k} &= \beta_1 \cdot m_{k-1} + (1-\beta_1)\cdot g_{k} \\
   v_{k} &= \beta_2 \cdot v_{k-1} + (1-\beta_2)\cdot g_{k}^2 \\
   \hat m_{k} &=  m_{k-1} /(1-\beta^k_1) \\
   \hat v_{k} &=  v_{k-1} /(1-\beta^k_2) \\
    d_k &= -\hat m_{k} /(\sqrt{\hat v_{k} }+\epsilon)
\end{split}
\label{eq:adamopt}
\end{equation}
Adam uses a momentum-driven strategy with a step-size correction mechanism based on the gradient variance. In the training of many architectures, especially Transformers, these changes have been shown to be important to produce high quality results \cite{chen22a}. To perform a weight update the general rule is:

\begin{equation}
     w_{k+1} = w_{k} + \eta_k d_k.
\end{equation}
The Armijo line search criterion from Eq. \ref{eq:armijo} must be adjusted for the Adam optimizer. We perform this adjustment based on \cite{vaswani20a, vaswani2021adaptive}.
To check if the Armijo criterion is satisfied in combination with Adam, the direction $d_k$ as defined in Eq. \ref{eq:adamopt} is used but with momentum $\beta_1 = 0$. Note that, the Armijo criterion is only guaranteed to be satisfy-able by adjusting the step size $\eta_k$, if the update direction and the gradient direction are identical. However, this condition is not met when $\beta_1 \neq 0$ in Eq. \ref{eq:adamopt}. 
Additionally, we replace the gradient norm term $||\nabla f_{k}(w_k)||^2$ by the preconditioned gradient norm $\frac{||\nabla f_{k}(w_k)||^2}{\sqrt{\hat v_{k} }+\epsilon}$ as in \cite{vaswani2021adaptive} resulting in Eq. \ref{eq:armijoadam}.
\begin{equation}
    f_{k}(w_k + \eta_k d_k) \leq f_{k}(w_k) - c \cdot \eta_k \frac{||\nabla f_{k}(w_k)||^2}{\sqrt{\hat v_{k} }+\epsilon}
    \label{eq:armijoadam}
\end{equation}

Note that to perform final weight updates each step  $\beta_1 \neq 0$ is used.

\subsection{Failure Cases}

As shown in \cite{vaswani20a, vaswani2021adaptive} the previously described line search methods perform well on smaller datasets and neural network architectures. However, here we show that these methods have problems to consistently perform during larger scale training. Especially on large scale transformer architectures which are notoriously sensitive to initial learning rates. 

\begin{figure}
    \includegraphics[width = 0.48\textwidth]{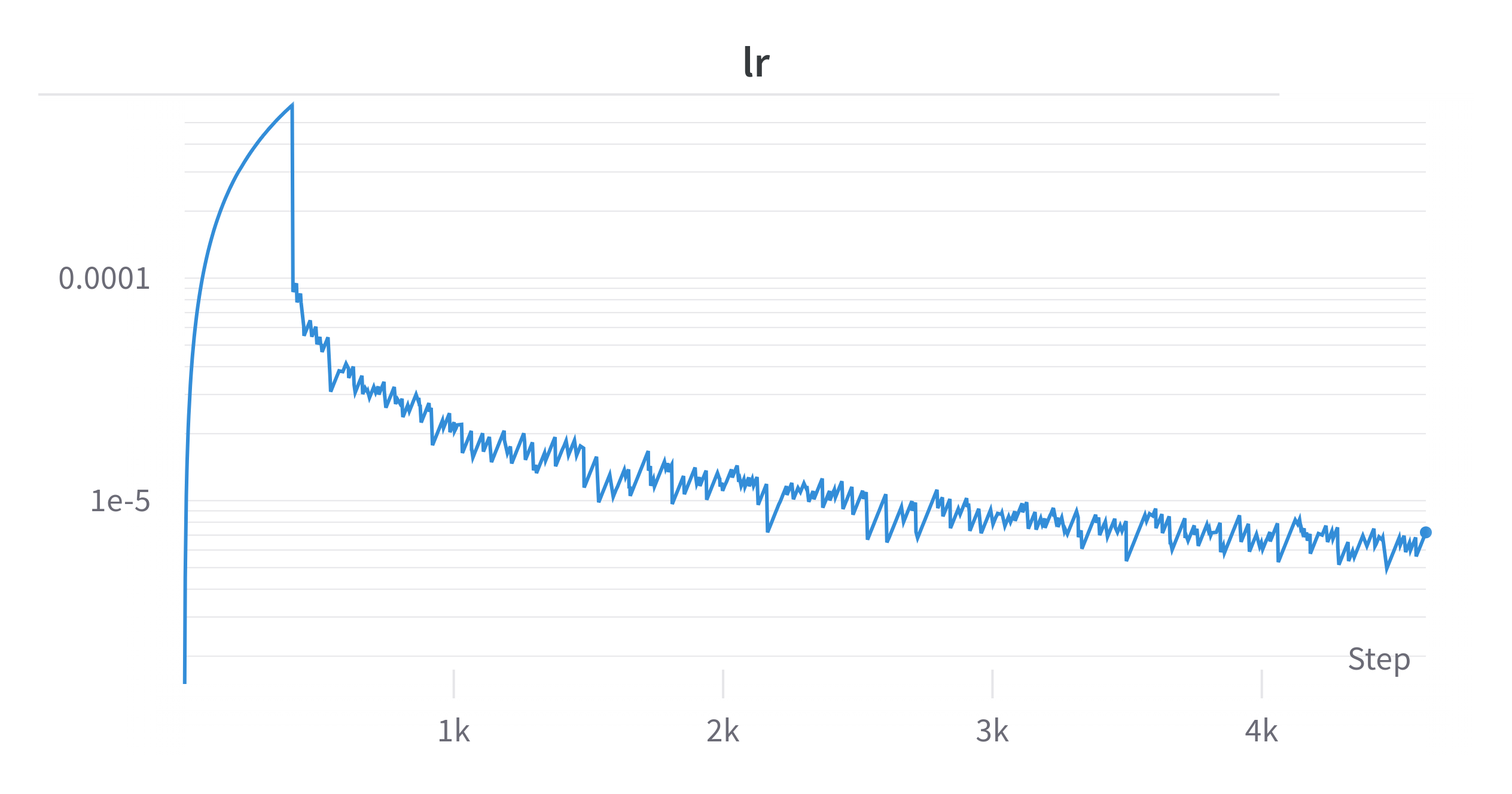}
  \caption{Step size $\eta_k$ for large scale GPT2 training. We started with a fixed linear warmup of the step size until step 400. Afterwards, ADAM + SLS determined the step size.}
\label{fig:step_sizegpt2}
\end{figure}

We identify one of the main causes for this discrepancy. We propose that the issue arises from the Armijo criterion exclusively conducting the line search in the direction of the gradient. When this direction significantly diverges from the actual update direction, as is often the case in large-scale transformer training, setting the momentum term $\beta_1 = 0$ becomes unreliable for estimating the optimal step size.
The resulting step size $\eta_k$ for large models trained on large scale data is too low in most cases, see Figure \ref{fig:step_sizegpt2}. Here we quickly converge to step sizes of in the range of  $[1e{-5},1e{-6}]$ where a appropriate step sizes would be in the range of $[1e{-3},5e{-5}]$


%% file: methods.tex
\label{sec:alsals}
To obtain a line search method with better properties, we propose to extend the Armijo criterion. Below we provide a detailed explanation of the modifications we made and our reasoning.

\subsection{Analyzing the Loss Landscape} 

In this section we take a closer look at typical loss landscapes in our experiments and the resulting step sizes.

\begin{figure}
    \includegraphics[width = 0.48\textwidth]{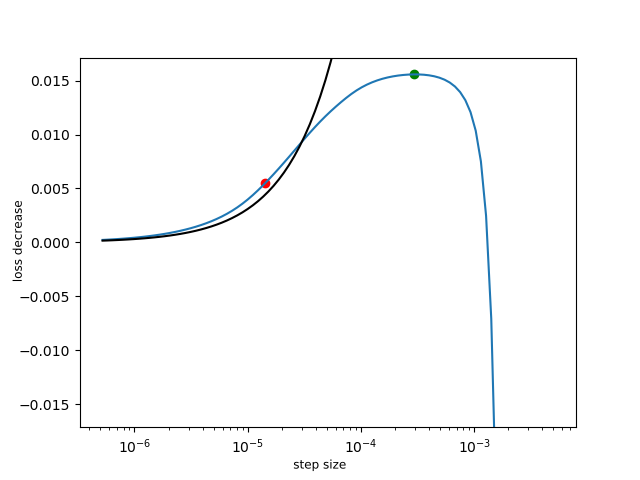}
  \caption{Loss decrease (y-axis) vs step size (x-axis) on the QNLI dataset for a single batch. Note the logarithmic scaling of the x-axis. Red point indicates step size resulting of ALSALS. Green point indicates optimum loss decrease on single batch. The Area above the black line indicates where Eq \ref{eq:armijoadam} is satisfied.}
\label{fig:losslssingle}
\end{figure}

In Figure \ref{fig:losslssingle}, we see a typical loss landscape. With increasing step size the loss decreases down to a single minimum, after which it sharply increases. 
This is to be expected as we view only a slice of the whole loss landscape, namely the gradient direction which is defined being the direction with the largest decrease in loss.

\begin{figure}
    \includegraphics[width = 0.48\textwidth]{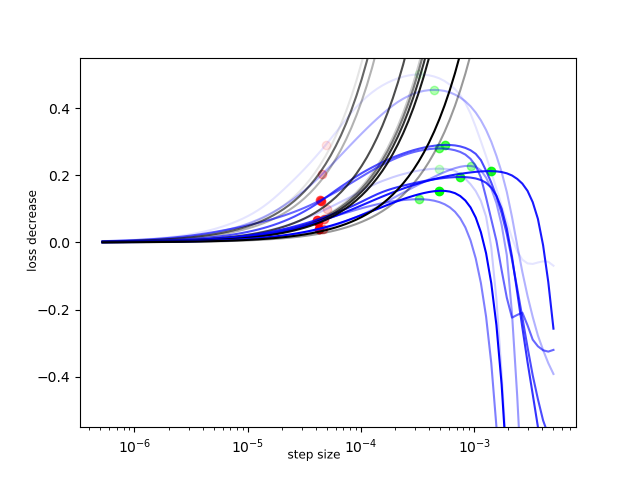}
  \caption{Loss decrease (y-axis) vs step size (x-axis) on QNLI training for the last 10 consecutive batches with older runs fading. Note the logarithmic scaling of the x-axis. Red points indicate step size resulting of ALSALS. Green points indicate optimum loss decrease on single batch. }
\label{fig:losslsmultiple}
\end{figure}

The main difference between different loss landscape plots we observe is the point at which the maximum loss decrease is located, see Figure \ref{fig:losslsmultiple}.

In practice it is important to always chose a step size which is below this maximum loss decrease of a single batch, since this varies highly over batches and one would risk otherwise entering the area of sharp loss increase over the whole dataset. To illustrate this point, we trained models by always selecting the optimum step size according to the full line search. This resulted in diverging runs, clearly showing that a conservative estimate is needed.

\begin{figure}
    \includegraphics[width = 0.48\textwidth]{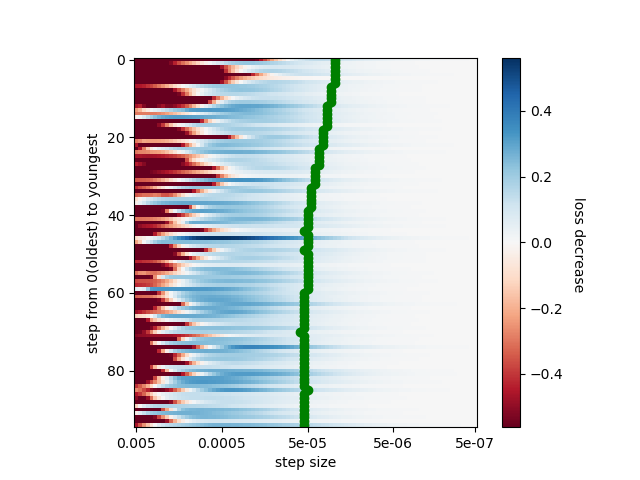}
  \caption{Loss decrease depicted as color on the QNLI dataset. Step $k$ is displayed on the y axis and step size $\eta_k$ on the x axis. Note the logarithmic scaling of the x-axis. Green line indicates step size resulting of ALSALS.}
\label{fig:losslslarge}
\end{figure}

Visualized in Figure \ref{fig:losslslarge} we plot the loss landscape and its changing nature via a height plot. Here we once again see the differences of the loss landscape between batches. 

\subsection{Addressing the momentum term of Adam} 

It is greatly desirable to perform the line search with $\beta_1$ set to its normal value. However, this leads to many problems with the current criterion from Equation \ref{eq:armijo}. In many cases for arbitrary $\eta \in [0,\infty]$ this criterion is not possible to be satisfied, resulting in the step size $\eta = 0$. 

We introduce a criterion which works while taking the momentum term of ADAM into account.
The main idea behind the new criterion is to approximate the change of loss when taking a step in the ADAM direction. 

Following this idea, in Equation \ref{eq:armijo}, we replace $||\nabla f_{k}(w_k)||^2$ with a term we call gradient magnitude approximation $f_a$ to be calculated as follows: 
\begin{equation}
    f_a = \frac{f_{k}(w_k) - f_{k}(w_k +  \cdot \lambda \cdot d_k)}{\epsilon}
\end{equation}
with $\epsilon$ being some small value in our case we choose $\epsilon = 5*10^{-8}$. 
$f_a$ is a close approximation of $||\nabla f_{k}(w_k)||^2$ for step directions $d_k$ without a momentum term, but critically does yield the loss decrease (or increase) for $d_k$ with an applied momentum term. This results in:
\begin{equation}
    f_{k}(w_k) - f_{k}(w_k + \eta_k d_k) \geq  c \cdot \eta_k \cdot f_a
    \label{eq:armijobeta}
\end{equation}

In the case that $f_a \leq 0$ we need to further modify the criterion, since it would otherwise send the step size $\eta_k$ to zero, as no matter how small the step size the loss always increases. This is a phenomenon that can not occur in the original Armijo formulation, since stepping in the gradient direction is guaranteed to result in a loss decrease for a sufficiently small step size. 

\begin{equation}
c_h = 
\begin{cases} 
c & \text{if } f_a > 0 \\
c^{-1} & \text{else } f_a \leq 0 
\end{cases}
\end{equation}

resulting in:

\begin{equation}
    f_{k}(w_k) - f_{k}(w_k + \eta_k d_k) \geq  c_h \cdot \eta_k \cdot f_a
    \label{eq:armijobeta}
\end{equation}




\subsection{Practical Considerations}

As we changed the approximation of $||\nabla f_{k}(w_k)||^2$ we need to perform hyperparameter tuning for the $c$ value anew. 
In our experiments we found good values for $c$ to be in the range $c \in [0.3,0.7]$.  
For all our experiments we used $c= 0.6$ (compared to $c=0.1$ for the original Armijo line search criterion). 

%% file: experiments.tex
In this section, we elaborate on our experimental design aimed at assessing the effectiveness of the optimization method we have proposed.
We utilize datasets, model implementations and weights from the Huggingface library, the pytorch datasets library and the nanoGPT \cite{nanoGPT} github repository. 

\subsection{Candidates} 
\label{sec:candidates}
A quick overview of all candidates we are evaluating can be seen below:

\begin{itemize}
    \item ADAM with tuned learning rate and learning rate schedule
    \item ADAM + SLS, see Section \ref{sec:adamarmijo} 
    \item ALSALS, see Section \ref{sec:alsals}
    \end{itemize}

For a baseline comparison, we assess the performance of the ADAM optimizer using a cosine decay strategy with warm starting applied for 10\% of the entire training duration.For NLP tasks this warm starting and cosine decay is common practice. For the image tasks we compare to a flat learning rate as done in \cite{vaswani20a}. 

We take the peak learning rate for ADAM on natural language tasks $\eta = 2 \cdot 10^{-5}$ from the original Bert paper \cite{bert}, which presents a good value for numerous classification tasks, encompassing the Glue tasks, which we assess in our evaluation.

For GPT-2 training, we use the peak learning rate of $\eta = 6 \cdot 10^{-4}$ as described in \cite{gpt} and use a warm-starting period of 2000 steps for all algorithms.

We found the value $\eta = 1 \cdot 10^{-3}$ for image classification for ADAM using a grid search. 

\subsection{Datasets and Models}
To evaluate an optimization method it is necessary to perform large scale runs of complex real world datasets and tasks. This is especially important as many optimization methods perform well on small scale or clean theoretical tasks, but fail to perform well on real world data.

\emph{Natural Language Processing - Transformers}
As the most important evaluation metric we use large scale transformer training. For a specific implementation we choose to train the GPT-2 Model \cite{gpt} on openwebtext \cite{Gokaslan2019OpenWeb}. We use the nanoGPT implementation \cite{nanoGPT}, which follows all best practices for large scale training.

Another common scenario in natural language processing is fine tuning a language model for example Bert \cite{bert}.
To evaluate this scenario we choose the Glue dataset \cite{wang-etal-2018-glue}. 

More specifically of the Glue collection \cite{wang-etal-2018-glue}, we use the datasets MRPC, SST2, MNLI and QNLI. These datasets range from 500 - 400.000 training samples and represent a variety of different fine-tuning tasks.

\emph{Image Classification - Convolutional Neural Networks}

In image classification common evaluation datasets are CIFAR10 and CIFAR100 \cite{cifar}, both being small scale (50.000 samples, 32x32 resolution). To obtain more reliable results we also compare on ImageNet \cite{imagenet} which consists of roughly $10^6$ samples which we scale to 224x224. 
We use the ResNet34 \cite{resnet} architecture for the CIFAR datasets and ResNet50 \cite{resnet} for ImageNet. A larger architecture is used for ImageNet due to the increased amount of complexity and size of the dataset.


\subsection{Implementation Details} 
The following details are the same for all experiments:
We train all models 5 times and the report the average metrics in Tables \ref{Fig:acc} and \ref{Fig:loss}. The learning curves as well as standard errors are visualized in Figures \ref{fig:gpt2loss},\ref{fig:nlp} and \ref{fig:image}. 

A Bert \cite{bert} model was trained on the NLP dataset with the following hyperparameter choices: Five epochs training time on each dataset. The Glue experiments employed the [CLS] token for the pooling operation. The maximum sequence length was configured to accommodate $256$ tokens, while a batch size of $32$ was utilized during the training process.

For the image datasets the batch size used during training is set to 128. We applied pre-processing as described in the ResNet paper \cite{resnet}. Models were trained on CIFAR10 and CIFAR100 for 100 epochs and on ImageNet for 5 epochs.



%% file: results.tex
\begin{figure*}
\vspace{-0.05\textwidth}
\subfloat[ImageNet]{\includegraphics[width = 0.33\textwidth]{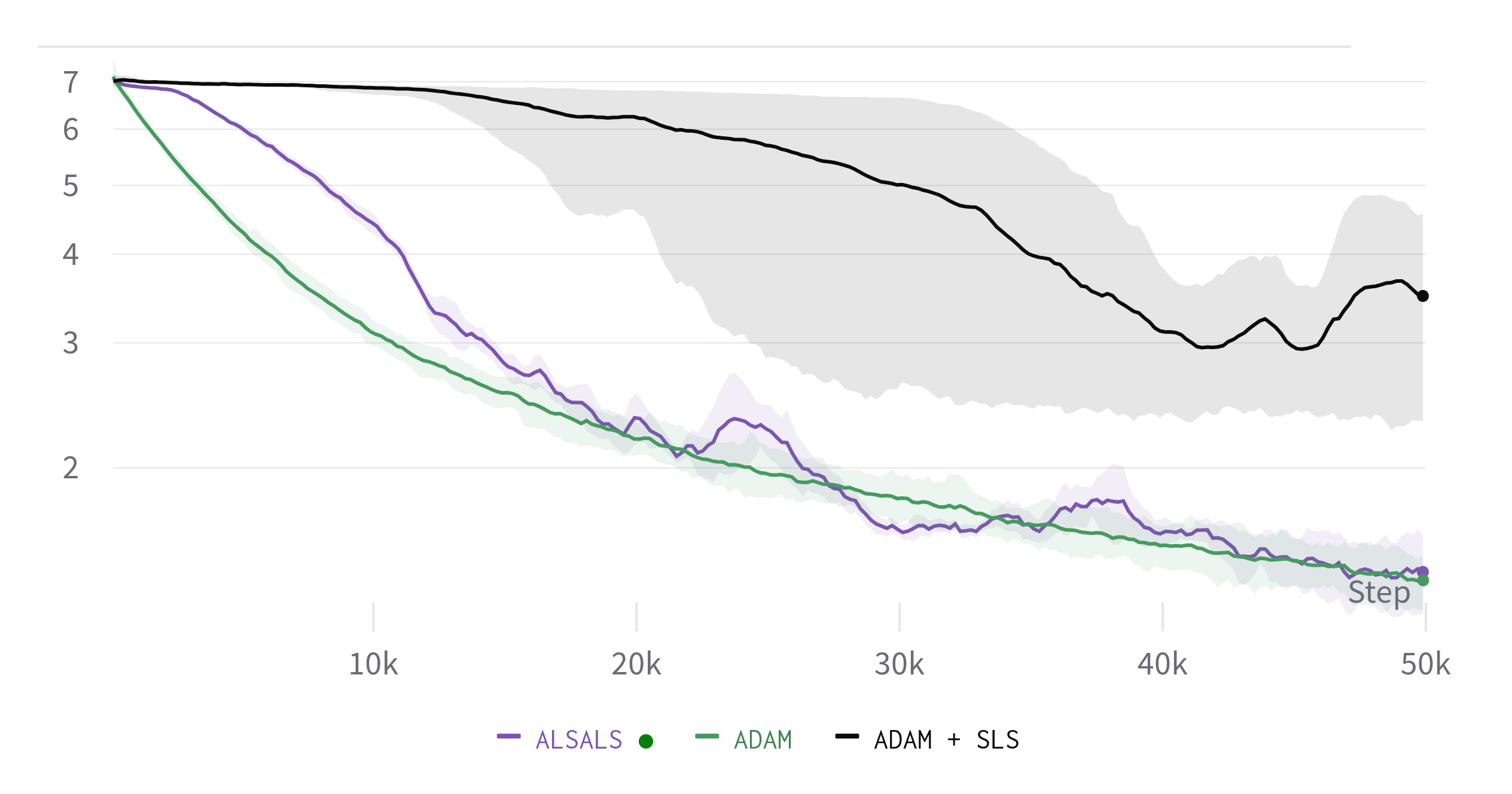}}
\subfloat[Cifar 100]{\includegraphics[width = 0.33\textwidth]{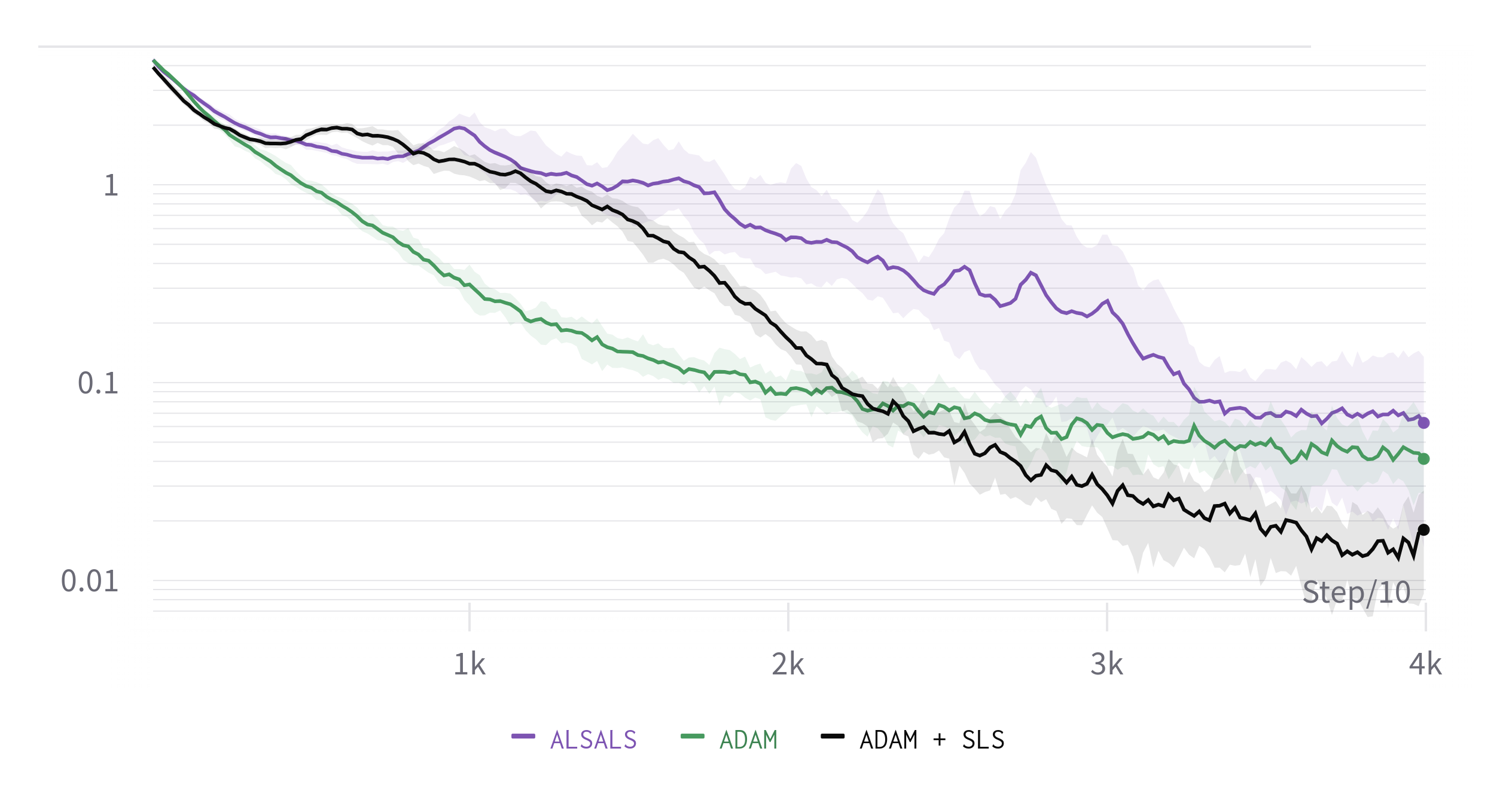}}
\subfloat[Cifar 10]{\includegraphics[width = 0.33\textwidth]{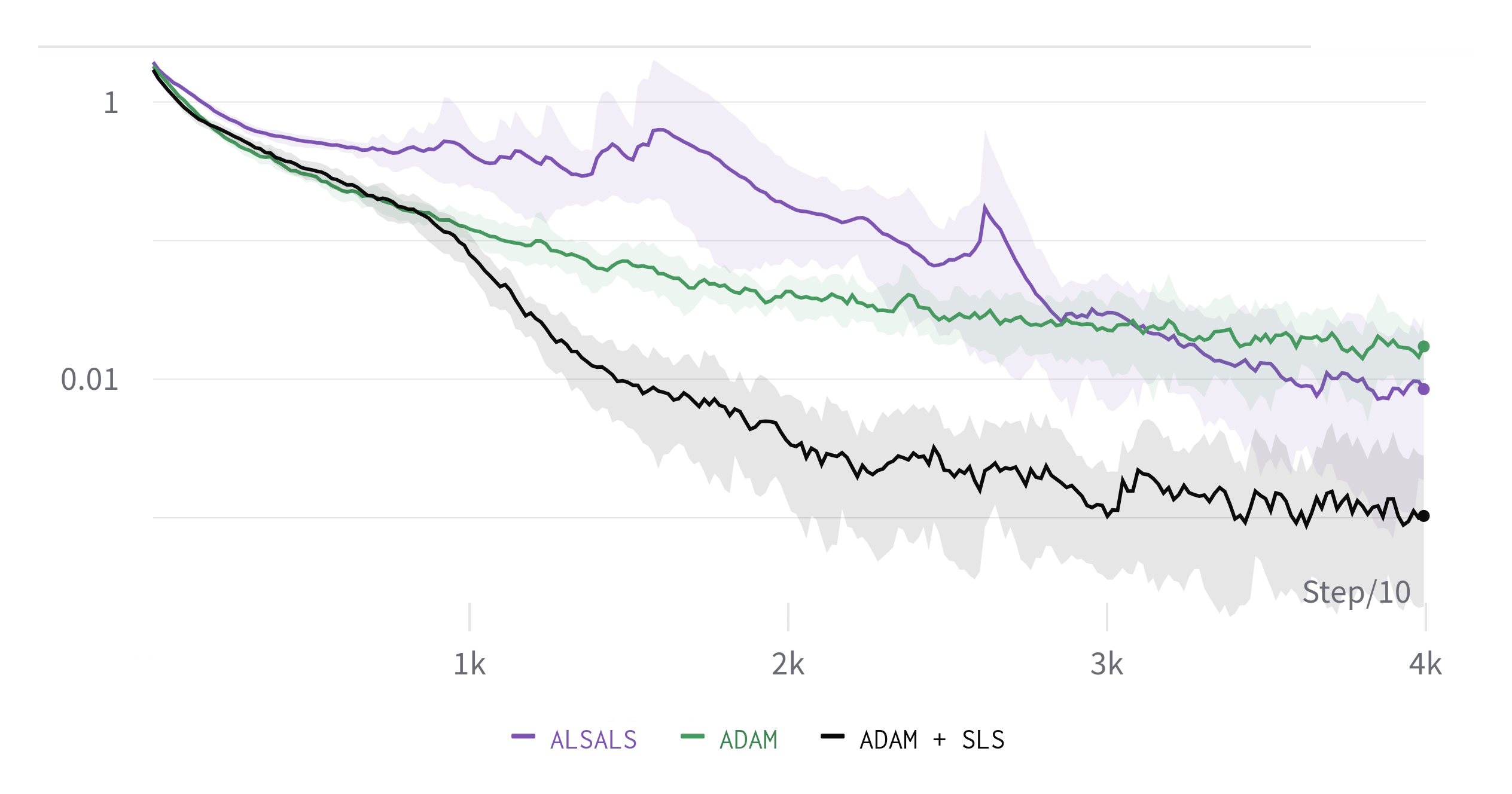}} \\
\subfloat[ImageNet]{\includegraphics[width = 0.33\textwidth]{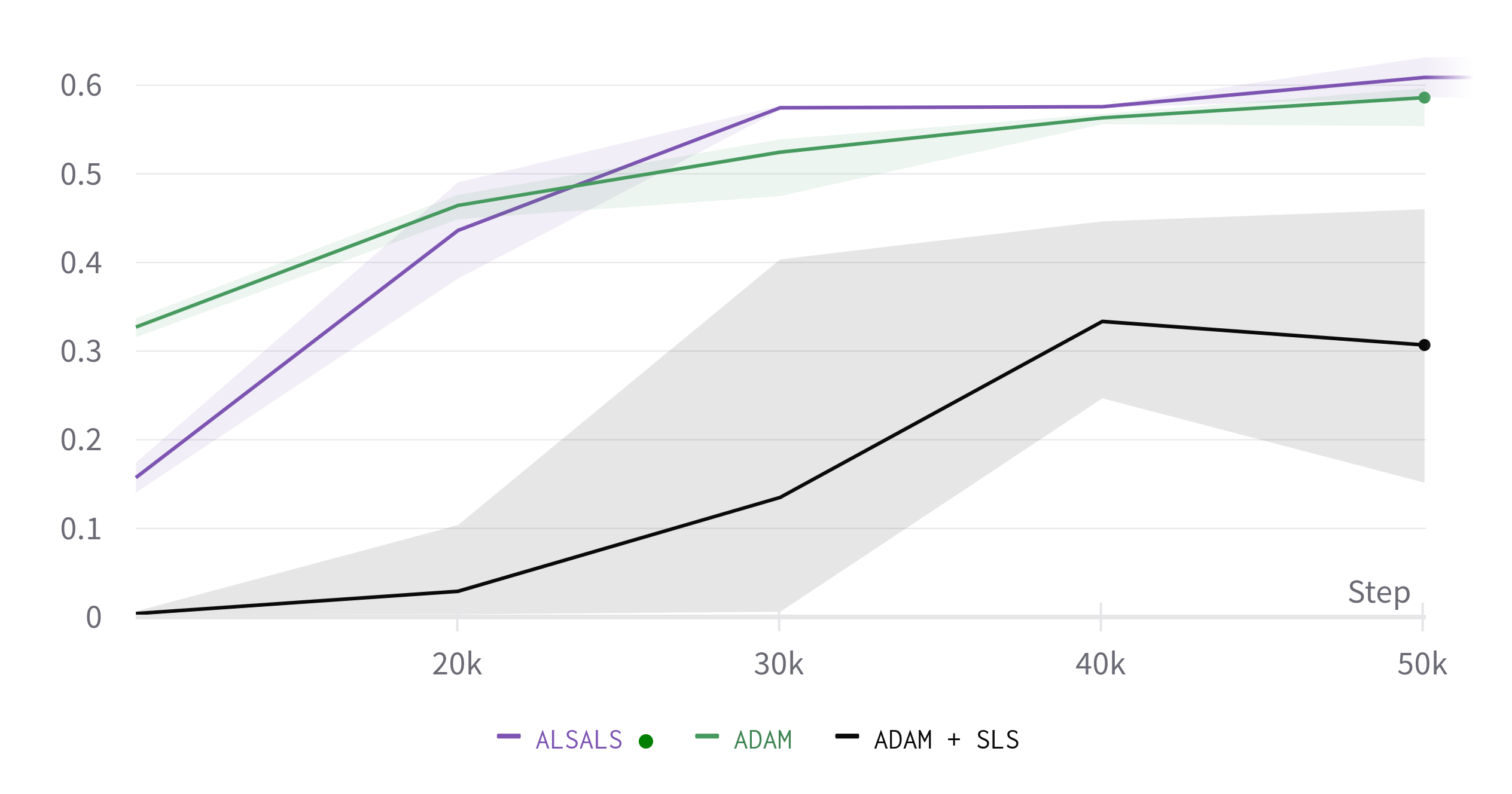}} 
\subfloat[Cifar 100]{\includegraphics[width = 0.33\textwidth]{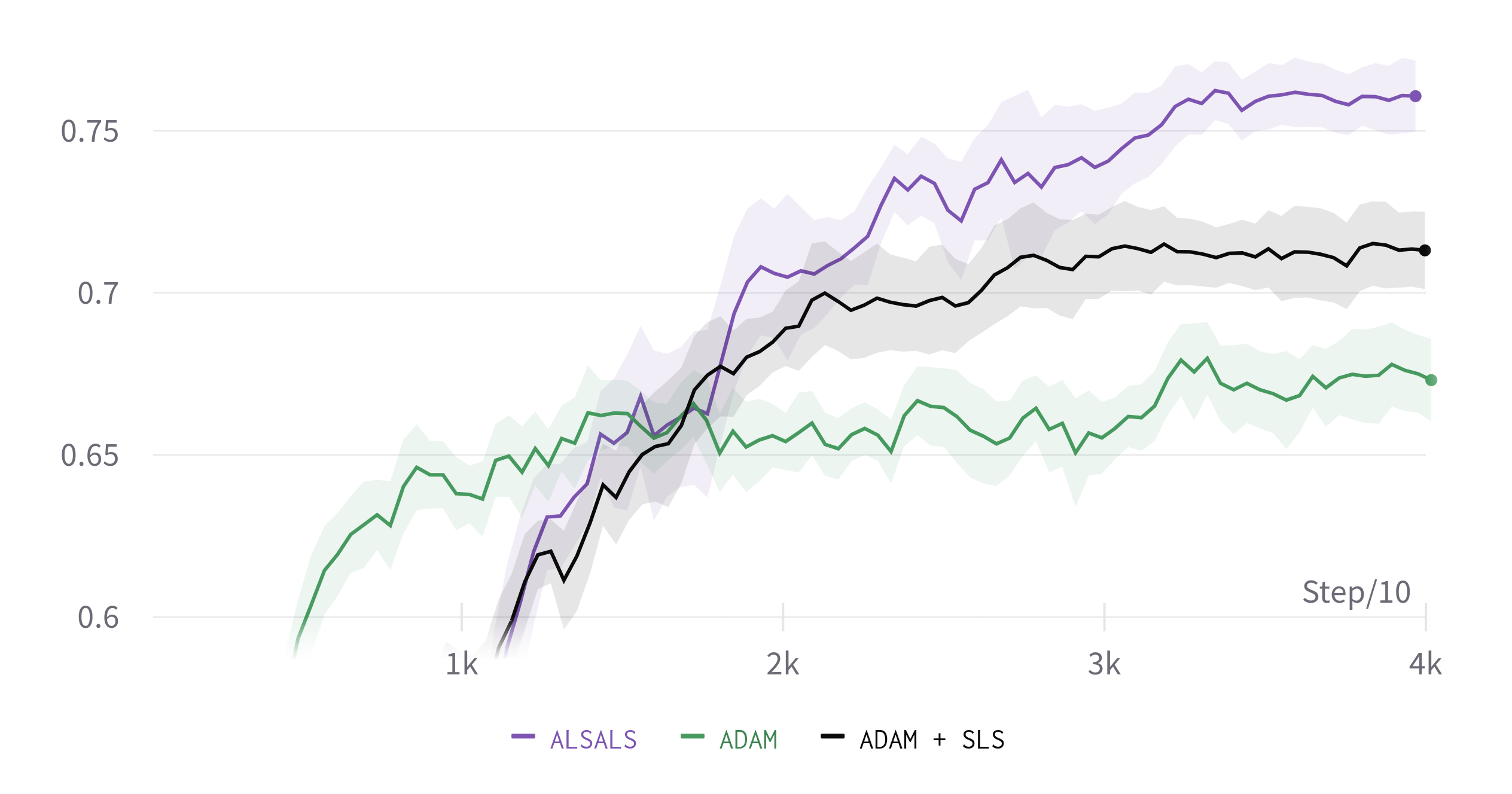}} 
\subfloat[Cifar 10]{\includegraphics[width = 0.33\textwidth]{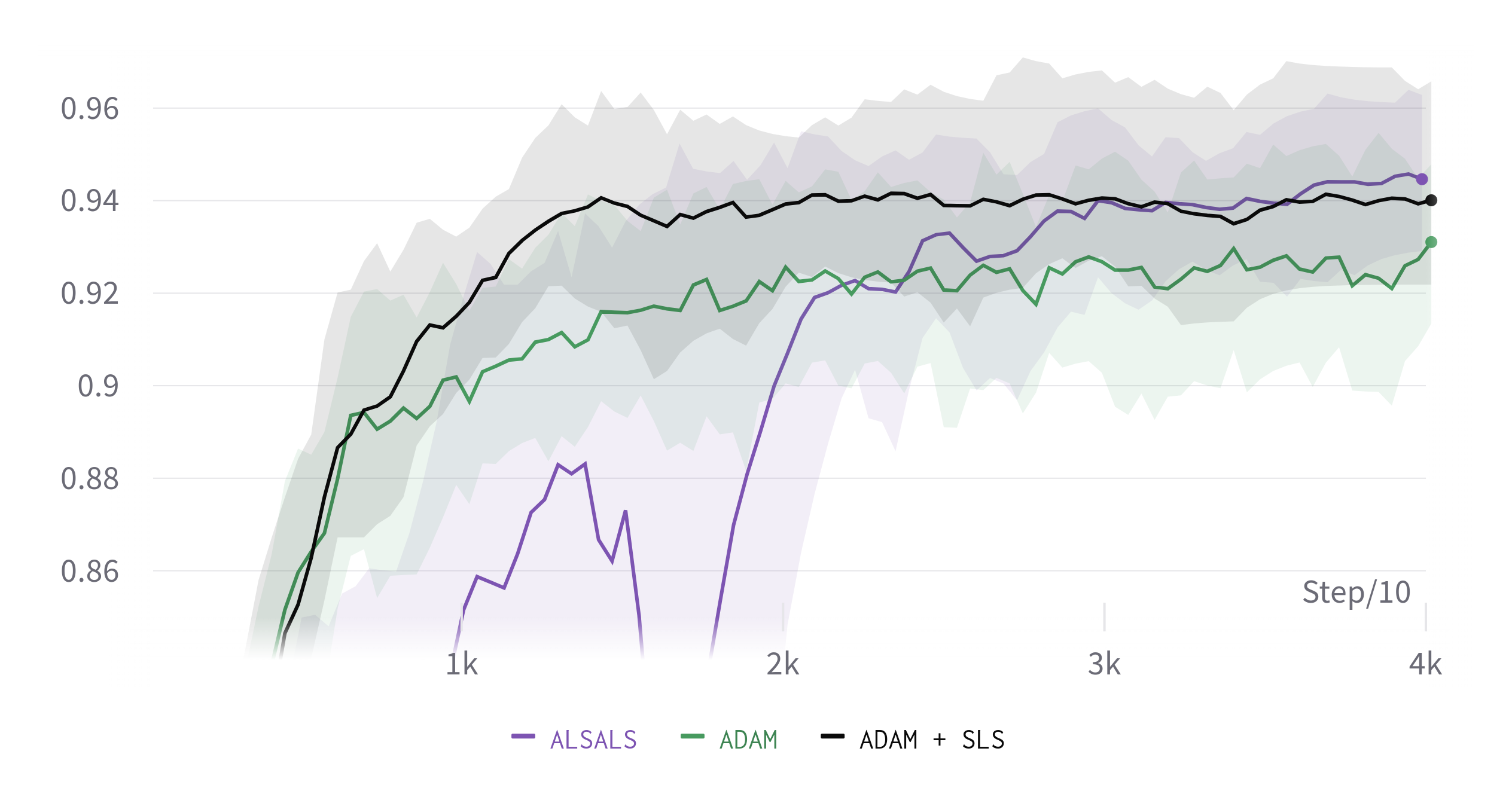}}

\caption{The top row displays the loss curves, while the bottom row presents the accuracy curves for the ResNet experiments on image datasets. Standard errors are indicated around each line, beginning from the second epoch. Accuracy was computed on the validation data, whereas loss was assessed on the training data.}
\label{fig:image}
\vspace{-0.01\textwidth}
\end{figure*}

In this section, we will show the results of our experiments. 
We compare the 3 candidates described in Section \ref{sec:candidates}. All metrics reported are average values obtained using 5 training runs.

All displayed accuracies are computed using validation sets. The losses presented are derived from the training sets and are smoothed using an exponential moving average. The shaded regions surrounding each line represent the standard error for each experiment. We present the accuracies and losses observed throughout the training period in Figures \ref{fig:gpt2loss}, \ref{fig:nlp}, and \ref{fig:image}. Tables \ref{Fig:acc} and \ref{Fig:loss} illustrate the peak accuracies and final loss for each candidate.

\subsection{Natural Language Processing - Transformer Experiments}


\begin{figure}
    \centering
    \includegraphics[width = 0.48\textwidth]{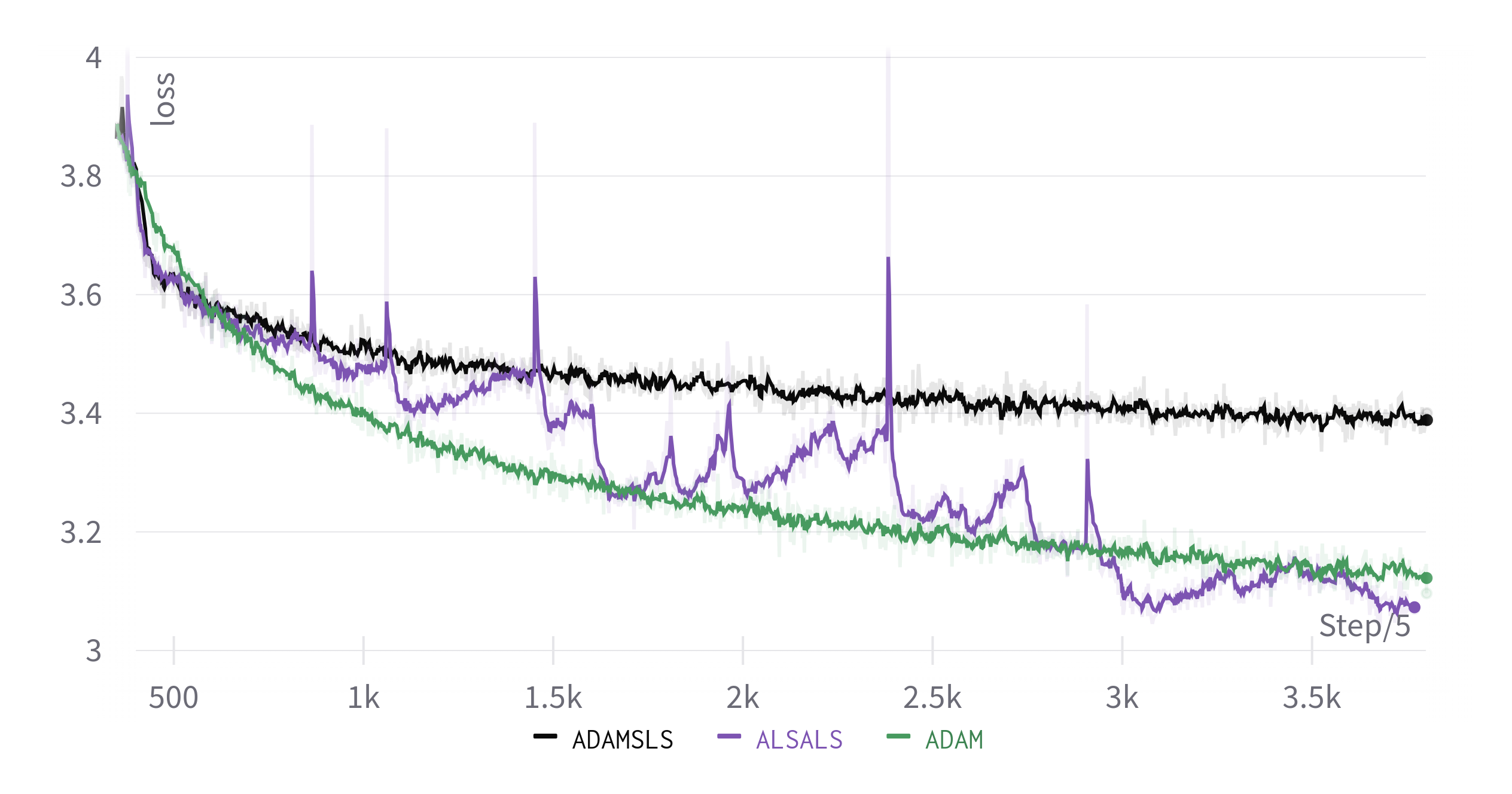}
    \caption{Training loss of GPT-2 during large scale training with different step size methods.}
    \label{fig:gpt2loss}
\end{figure}

\begin{figure*}
\subfloat[MNLI ]{\includegraphics[width = 0.33\textwidth]{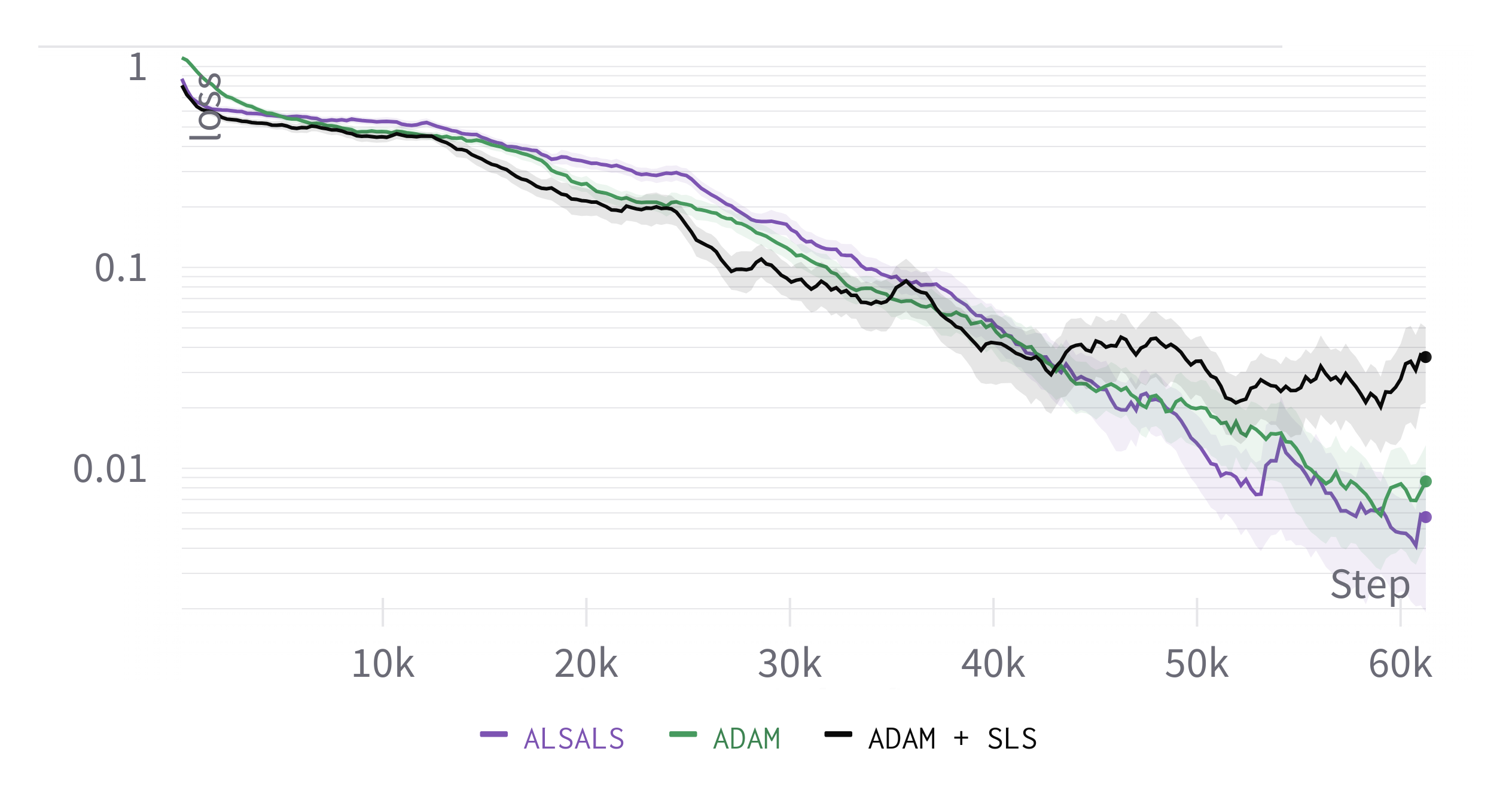}} 
\subfloat[MRPC ]{\includegraphics[width = 0.33\textwidth]{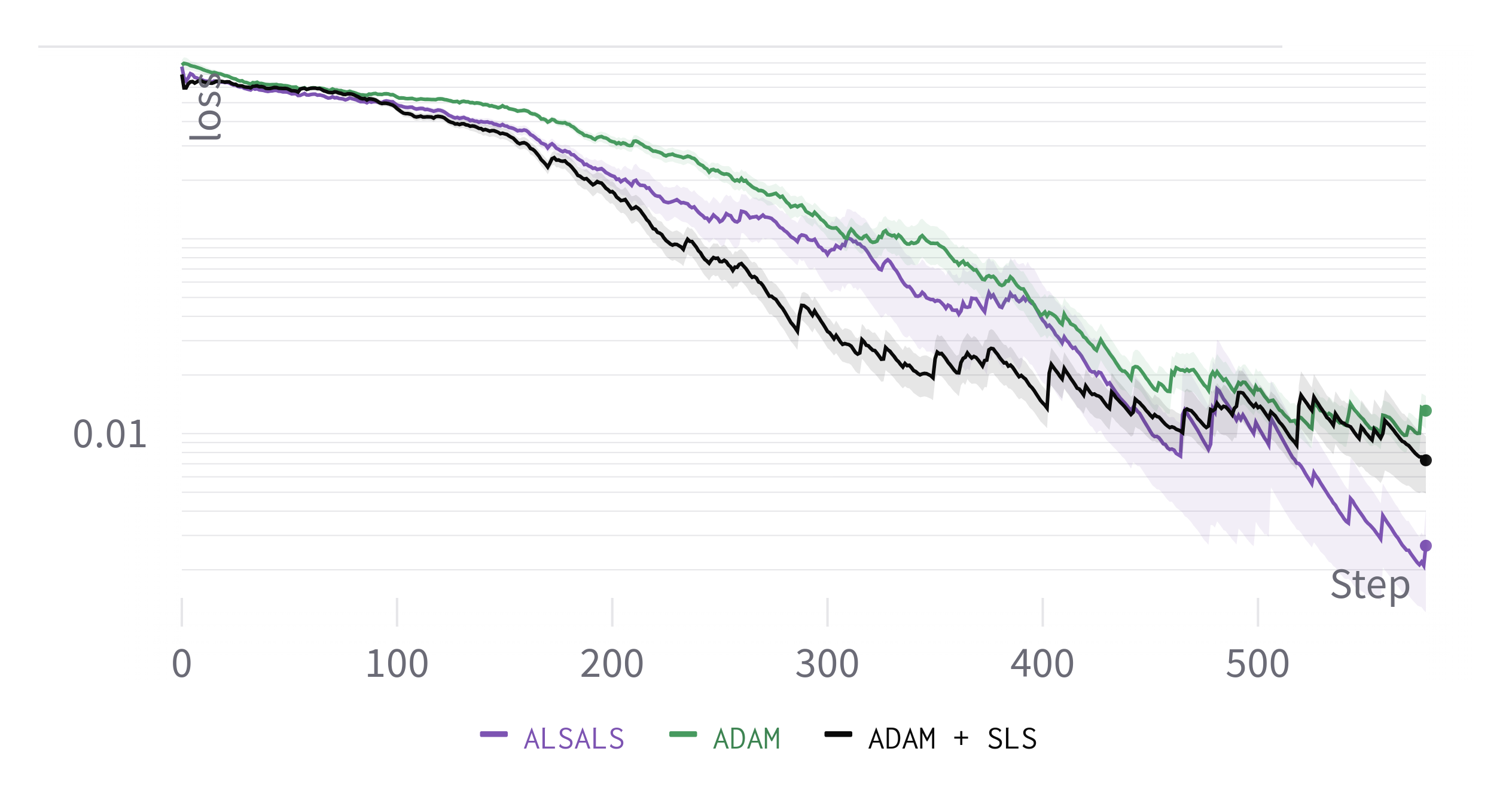}}
\subfloat[QNLI ]{\includegraphics[width = 0.33\textwidth]{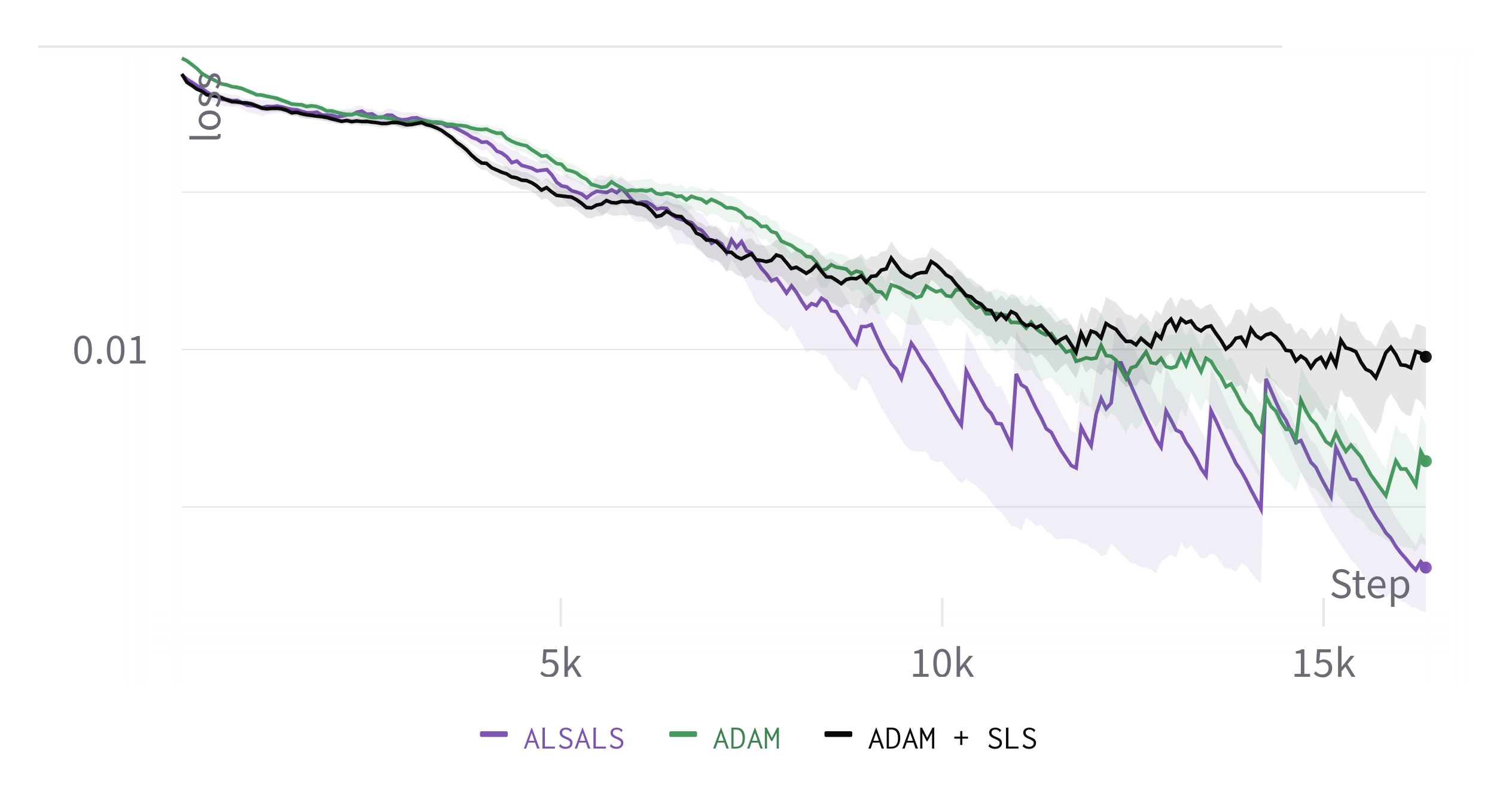}}
 \\
\subfloat[MNLI]{\includegraphics[width = 0.33\textwidth]{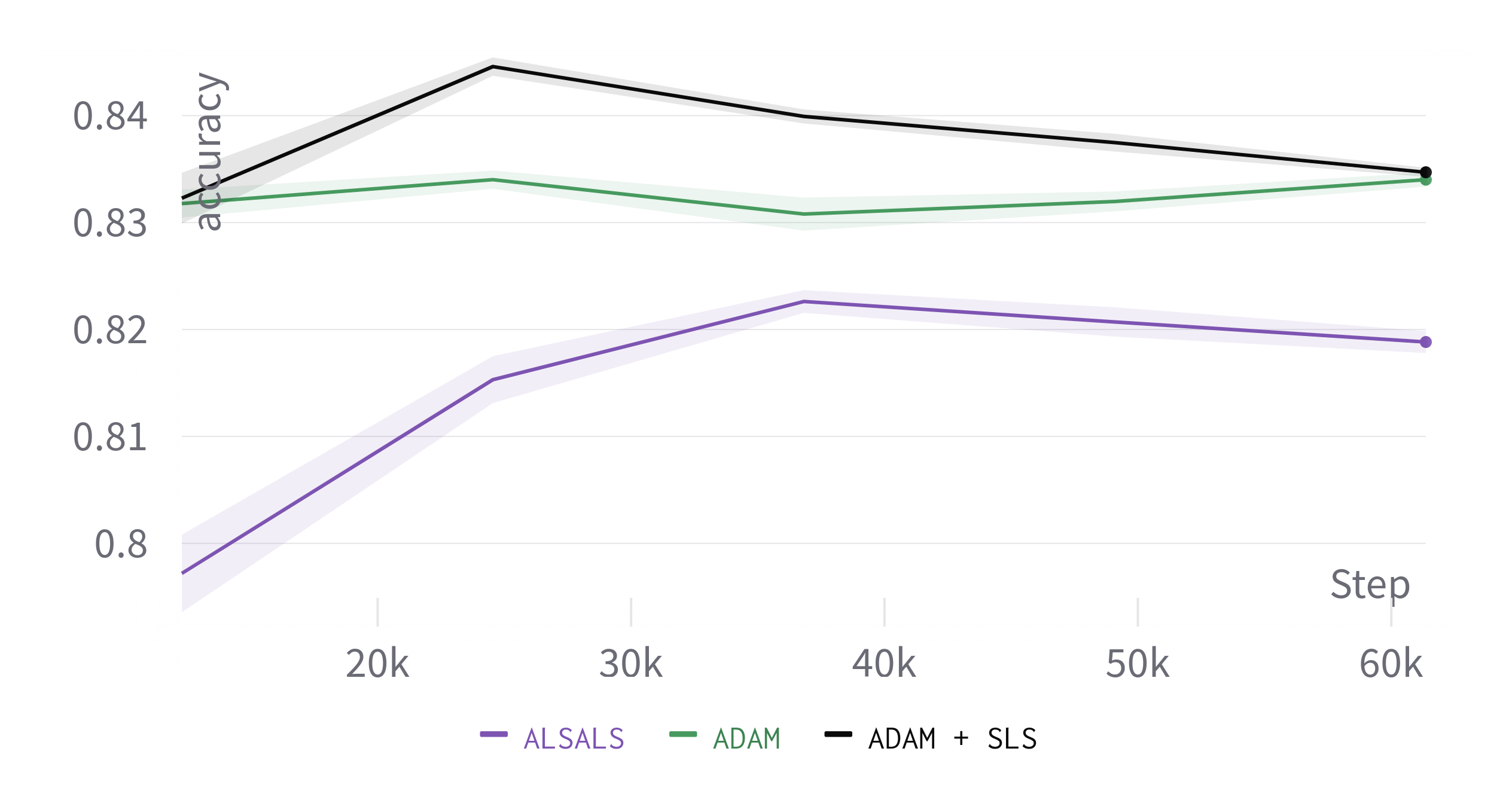}} 
\subfloat[MRPC]{\includegraphics[width = 0.33\textwidth]{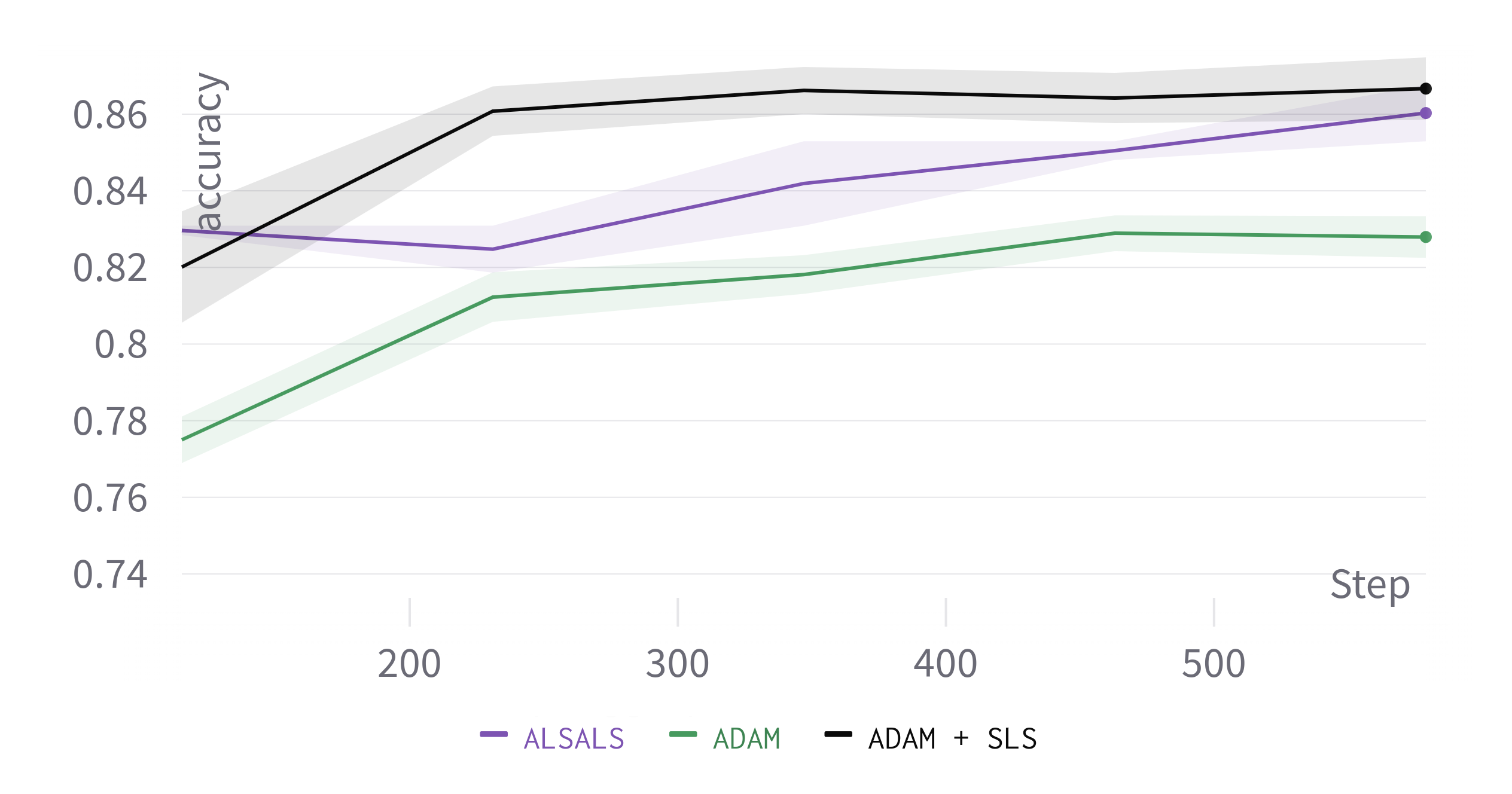}}
\subfloat[QNLI]{\includegraphics[width = 0.33\textwidth]{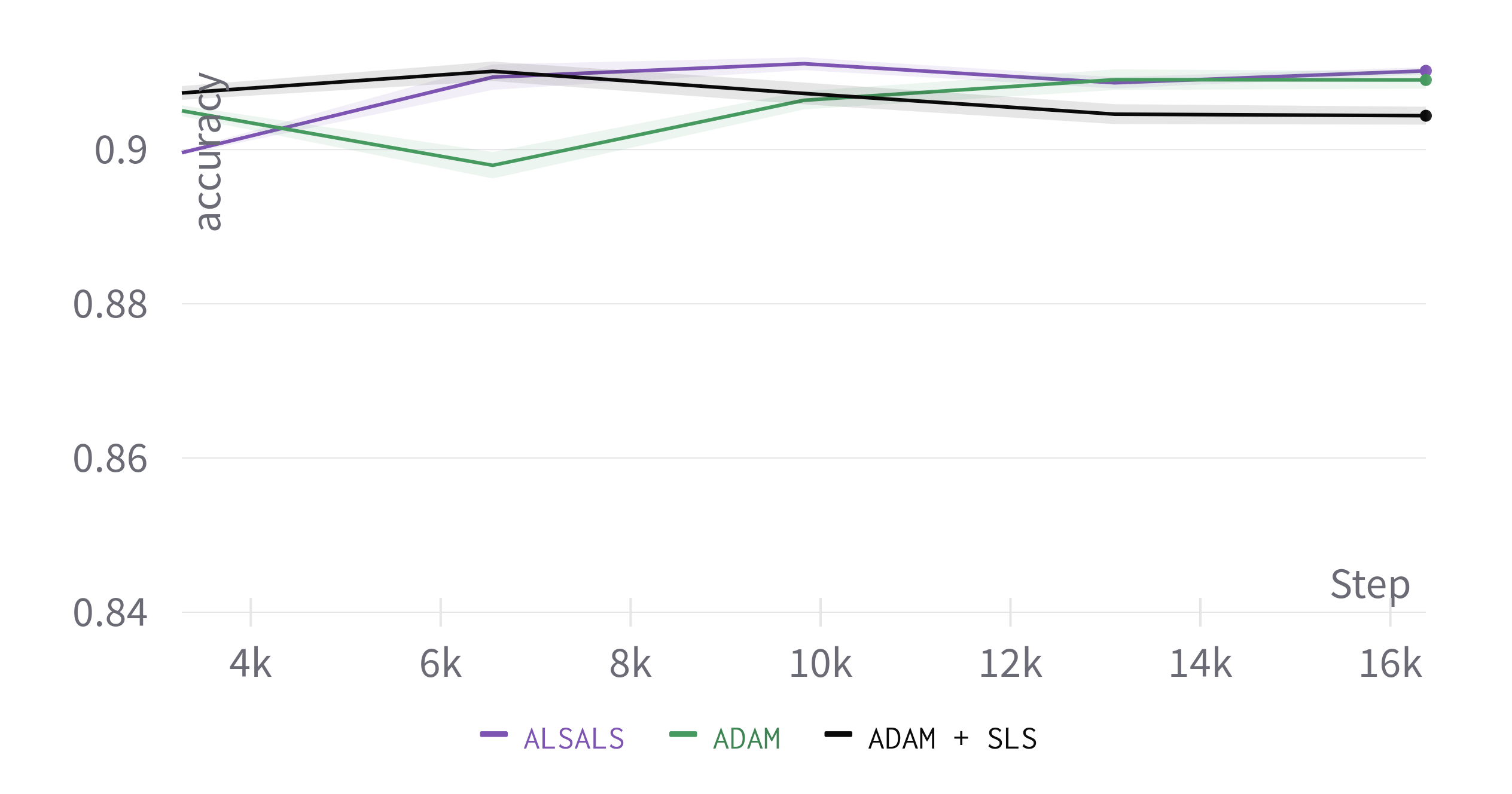}}

\caption{
The top row displays the loss curves, while the bottom row presents the accuracy curves for the experiments on the GLUE datasets. Standard errors are indicated around each line, beginning from the second epoch. Accuracy was computed on the validation data, whereas loss was assessed on the training data. }
\label{fig:nlp}
\end{figure*}

In our NLP experiments, as shown in Figure \ref{fig:gpt2loss}, \ref{fig:nlp} and Table \ref{Fig:loss}, \ref{Fig:acc}, we have observed that, on average, ALSALS achieves a lower final loss compared to ADAM or ADAM + SLS. However, this improvement in loss does not always translate to a significant difference in the accuracy metric.

\begin{table}[t] 
  \centering
  \caption{Peak classification accuracies, averaged over 5 runs, for all datasets and optimization methods. Best performing optimization method is marked in \textbf{bold}. }
  \label{Fig:acc}
  \begin{tabular}{c ccc}
    \toprule
 & ADAM &  ADAM + SLS & ALSALS \\
 \cmidrule(r){1-1}   \cmidrule(r){2-4} 
\it{MNLI} & 0.8340  & \textbf{0.8347} & 0.8188    \\
\it{QNLI} & 0.9090 & 0.9044  & \textbf{0.9102}    \\
\it{MRPC} & 0.8279 & \textbf{0.8667}  & 0.8603   \\
\it{SST2} & \textbf{0.9271} & 0.9261   & 0.9128   \\
 \cmidrule(r){1-1}   \cmidrule(r){2-4} 

 ResNet34 \\
\it{CIFAR10} & 0.9273 & 0.9393 & \textbf{0.9446}    \\
\it{CIFAR100} & 0.675 &  0.7131 & \textbf{0.7607}   \\
 ResNet50 \\
\it{ImageNet} & 0.5860 &  0.3069 & \textbf{0.6314}    \\
 \cmidrule(r){1-1}   \cmidrule(r){2-4} 

average & 0.8123 & 0.7844 & \textbf{0.8341}     \\
    \bottomrule
  \end{tabular}
\end{table}

\begin{table}
  \centering
  \caption{Final losses, averaged over 5 runs, for all datasets and optimization methods. Best performing (minimal loss) optimization method is marked in \textbf{bold}. The logarithmic average is taken due to the logarithmic nature of the typical loss.}
  \label{Fig:loss}
  \begin{tabular}{c ccc}
    \toprule
 & ADAM &  ADAM + SLS & ALSALS \\
 \cmidrule(r){1-1}   \cmidrule(r){2-4} 
\it{MNLI} & 0.008607 & 0.0358  & \textbf{0.005717} \\
\it{QNLI} & 0.001953 &  0.008987  & \textbf{0.004116} \\
\it{MRPC} & 0.01312 & 0.007298  & \textbf{0.002657} \\
\it{SST2} & \textbf{0.006157} & 0.00822   & 0.01017  \\
GPT-2 & 3.135 & 3.395 & \textbf{3.078} \\
 \cmidrule(r){1-1}   \cmidrule(r){2-4} 
ResNet34\\
\it{CIFAR10} & 0.01725 &  \textbf{0.001032} & 0.008475   \\
\it{CIFAR100} & 0.04116 &  \textbf{0.01803}  & 0.06258  \\
ResNet50\\
\it{ImageNet} & 1.388 &  3.493 & \textbf{1.324}  \\
 \cmidrule(r){1-1}   \cmidrule(r){2-4} 

log average & 0.03783 &  0.03479  & \textbf{0.02467}  \\
    \bottomrule
  \end{tabular}
\end{table}

\subsection{Image - Convolutional Neural Networks Experiments}

In our image experiments, see Figure \ref{fig:image} and Table \ref{Fig:acc}, \ref{Fig:loss}, we have observed that ALSALS does perform the best on accuracy. ADAM + SLS performs better on the loss metric for CIFAR10 and CIFAR100, however it fails to perform well for any metric on ImageNet.

Although we do not observe a monotonically decreasing loss during training we converge consistently and observe better final performance on loss and accuracy. 




%% file: conclusion.tex

We have introduced ALSALS, an automatic step size selection method and built a hyperparameter free general purpose optimizer on top. We have compared its performance against tuned learning rates for larger datasets and architectures than previously done in optimizer evaluations for line search methods. The ALSALS optimizer performance compares favorably in these cases, while requiring no tuning of learning rates, or code overhead, as well as minimal compute overhead. Furthermore ALSALS is the first Line Search method we know capable of training large scale architectures, which was previously not possible with these methods. We recommend its use as a first choice for tuning deep neural networks in these domains and publish the code as a Python package \url{https://github.com/TheMody/Improving-Line-Search-Methods-for-Large-Scale-Neural-Network-Training}.






%% file: main.bbl
\begin{thebibliography}{10}

\bibitem{robbins51a}
H.~Robbins and S.~Monro, ``A stochastic approximation method,'' {\em The annals of mathematical statistics}, pp.~400--407, 1951.

\bibitem{DBLP:journals/corr/abs-1908-03265}
L.~Liu, H.~Jiang, P.~He, W.~Chen, X.~Liu, J.~Gao, and J.~Han, ``On the variance of the adaptive learning rate and beyond,'' in {\em International Conference on Learning Representations}, 2020.

\bibitem{adamw}
I.~Loshchilov and F.~Hutter, ``Decoupled weight decay regularization,'' in {\em International Conference on Learning Representations}, 2019.

\bibitem{RMSprop}
G.~Hinton and N.~S.~K. Swersky, ``Lecture notes neural networks for machine learning,'' 2014.

\bibitem{adam}
D.~P. Kingma and J.~Ba, ``Adam: {A} method for stochastic optimization,'' in {\em 3rd International Conference on Learning Representations, {ICLR} 2015, San Diego, CA, USA, May 7-9, 2015, Conference Track Proceedings} (Y.~Bengio and Y.~LeCun, eds.), 2015.

\bibitem{schmidt2021descending}
R.~M. Schmidt, F.~Schneider, and P.~Hennig, ``Descending through a crowded valley-benchmarking deep learning optimizers,'' in {\em International Conference on Machine Learning}, pp.~9367--9376, PMLR, 2021.

\bibitem{vaswani20a}
S.~Vaswani, A.~Mishkin, I.~Laradji, M.~Schmidt, G.~Gidel, and S.~Lacoste-Julien, ``Painless stochastic gradient: Interpolation, line-search, and convergence rates,'' {\em NIPS'19: Proceedings of the 33rd International Conference on Neural Information Processing Systems}, 2019.

\bibitem{mahsereci15a}
M.~Mahsereci and P.~Hennig, ``Probabilistic line searches for stochastic optimization,'' {\em Advances in neural information processing systems}, vol.~28, 2015.

\bibitem{bollapragada18a}
R.~Bollapragada, J.~Nocedal, D.~Mudigere, H.-J. Shi, and P.~T.~P. Tang, ``A progressive batching l-bfgs method for machine learning,'' in {\em International Conference on Machine Learning}, pp.~620--629, PMLR, 2018.

\bibitem{paquette20a}
C.~Paquette and K.~Scheinberg, ``A stochastic line search method with expected complexity analysis,'' {\em SIAM Journal on Optimization}, vol.~30, no.~1, pp.~349--376, 2020.

\bibitem{vaswani2021adaptive}
S.~Vaswani, I.~H. Laradji, F.~Kunstner, S.~Y. Meng, M.~Schmidt, and S.~Lacoste-Julien, ``Adaptive gradient methods converge faster with over-parameterization (and you can do a line-search),'' 2021.

\bibitem{ijcnn2023}
P.~Kenneweg, L.~Galli, T.~Kenneweg, and B.~Hammer, ``Faster convergence for transformer fine-tuning with line search methods,'' in {\em 2023 International Joint Conference on Neural Networks (IJCNN)}, pp.~1--8, 2023.

\bibitem{armijo66a}
L.~Armijo, ``Minimization of functions having lipschitz continuous first partial derivatives,'' {\em Pacific Journal of mathematics}, vol.~16, no.~1, pp.~1--3, 1966.

\bibitem{chen22a}
F.~Kunstner, J.~Chen, J.~W. Lavington, and M.~Schmidt, ``Noise is not the main factor behind the gap between sgd and adam on transformers, but sign descent might be,'' in {\em The Eleventh International Conference on Learning Representations}, 2023.

\bibitem{nanoGPT}
A.~Karpathy, ``nanogpt.'' \url{https://github.com/karpathy/nanoGPT}, 2023.

\bibitem{bert}
J.~Devlin, M.-W. Chang, K.~Lee, and K.~Toutanova, ``Bert: Pre-training of deep bidirectional transformers for language understanding,'' 2019.

\bibitem{gpt}
A.~Radford, J.~Wu, R.~Child, D.~Luan, D.~Amodei, I.~Sutskever, {\em et~al.}, ``Language models are unsupervised multitask learners,'' {\em OpenAI blog}, vol.~1, no.~8, p.~9, 2019.

\bibitem{Gokaslan2019OpenWeb}
A.~Gokaslan, V.~Cohen, E.~Pavlick, and S.~Tellex, ``Openwebtext corpus.'' \url{http://Skylion007.github.io/OpenWebTextCorpus}, 2019.

\bibitem{wang-etal-2018-glue}
A.~Wang, A.~Singh, J.~Michael, F.~Hill, O.~Levy, and S.~Bowman, ``{GLUE}: A multi-task benchmark and analysis platform for natural language understanding,'' in {\em Proceedings of the 2018 {EMNLP} Workshop {B}lackbox{NLP}: Analyzing and Interpreting Neural Networks for {NLP}}, (Brussels, Belgium), pp.~353--355, Association for Computational Linguistics, Nov. 2018.

\bibitem{cifar}
A.~Krizhevsky, ``Learning multiple layers of features from tiny images,'' pp.~32--33, 2009.

\bibitem{imagenet}
J.~Deng, W.~Dong, R.~Socher, L.-J. Li, K.~Li, and L.~Fei-Fei, ``Imagenet: A large-scale hierarchical image database,'' in {\em 2009 IEEE Conference on Computer Vision and Pattern Recognition}, pp.~248--255, 2009.

\bibitem{resnet}
K.~He, X.~Zhang, S.~Ren, and J.~Sun, ``Deep residual learning for image recognition,'' in {\em 2016 IEEE Conference on Computer Vision and Pattern Recognition (CVPR)}, pp.~770--778, 2016.

\end{thebibliography}
